\documentclass[11pt]{article}
\usepackage{coling2016}
\usepackage{times}
\usepackage{url}
\usepackage{latexsym}

\title{CogALex-V Shared Task: ROOT18}

\author{Emmanuele Chersoni \\
  Aix-Marseille University \\
  {\tt emmanuelechersoni@gmail.com} \\\And
  Giulia Rambelli \\
  University of Pisa \\
  {\tt rambelligiulia@gmail.com} \\\AND
  Enrico Santus \\
  The Hong Kong Polytechnic University \\
  {\tt esantus@gmail.com}} 

\date{}

\begin{document}
\maketitle
\begin{abstract}
  In this paper, we describe ROOT 18, a classifier using the scores of several \textit{unsupervised distributional measures} as features to discriminate between semantically related and unrelated words, and then to classify the related pairs according to their semantic relation (i.e. \textit{synonymy}, \textit{antonymy}, \textit{hypernymy}, \textit{part-whole meronymy}). Our classifier participated in the CogALex-V Shared Task, showing a solid performance on the first subtask, but a poor performance on the second subtask. The low scores reported on the second subtask suggest that distributional measures are not sufficient to discriminate between multiple semantic relations at once.
\end{abstract}

\section{Introduction}

\blfootnote{
    %
    % for review submission
    %
    %\hspace{-0.65cm}  % space normally used by the marker
    %Place licence statement here for the camera-ready version, see
    %Section~\ref{licence} of the instructions for preparing a
    %manuscript.
    %
     %final paper: en-uk version 
    %
    % \hspace{-0.65cm}  % space normally used by the marker
     This work is licenced under a Creative Commons 
     Attribution 4.0 International Licence.
     Licence details:
     \url{http://creativecommons.org/licenses/by/4.0/}
    % 
    % % final paper: en-us version 
    %
    % \hspace{-0.65cm}  % space normally used by the marker
    % This work is licensed under a Creative Commons 
    % Attribution 4.0 International License.
    % License details:
    % \url{http://creativecommons.org/licenses/by/4.0/}
}

The system described in this paper has been designed for the CogALex-V Shared Task, focusing on the corpus-based identification of semantic relations. Since Distributional Semantic Models (henceforth DSMs) were proposed as a special topic of interest for the current edition of the CogALex workshop, we decided to base our classifier on a number of distributional measures that have been used by past Natural Language Processing (NLP) research to discriminate between a specific semantic relation and other relation types.

The task is splitted into the following subtasks:
\begin{itemize}
\item for each word pair, the participating systems have to decide whether the terms are semantically related or not (TRUE and FALSE are the only possible outcomes);
\item for each word pair, the participating systems have to decide which semantic relation holds between the terms of the pair. The five possible semantic relations are synonymy (SYN), antonymy (ANT), hypernymy (HYPER), meronymy (PART\_OF) and no semantic relation at all (RANDOM).
\end{itemize}
Our system managed to achieve good results in discriminating between related and random pairs in the first subtask, but unfortunately it struggled in the second one, also due to the high difficulty of the task itself. In particular, the recall for some of the semantic relations of interest seems to be extremely low, suggesting that our unsupervised distributional measures do not provide sufficient information to characterize them, and that it could be probably useful to integrate such scores with other sources of evidence (e.g. information on lexical patterns of word co-occurrence). 

The paper is organized as follows: in section 2, we summarize related works on the task of semantic relation identification; in section 3, we introduce our system, by describing the classifier and the features. Finally, in section 4 we present and discuss our results.

\section{The Task: Related Work}

Distinguishing between related and unrelated words and, then, discriminating among semantic relations are very important tasks in NLP, and they have a wide range of applications, such as textual entailment, text summarization, sentiment analysis, ontology learning, and so on. For this reason, several systems over the last few years have been proposed to tackle this problem, using both unsupervised and supervised approaches (see the works of Lenci and Benotto \shortcite{lenci2012identifying} and Shwartz et al. \shortcite{shwartz2016improving} on hypernymy; Weeds et al. \shortcite{weeds2014learning} and Santus et al. \shortcite{santus2016nine} on hypernymy and co-hyponymy; Mohammad et al. \shortcite{Mohammad13} and Santus et al. \shortcite{santus2014taking} on antonymy). However, many of these works focus on a single semantic relation, e.g. antonymy, and describe methods or measures to set it apart from other relations. There have not been many attempts, at the best of our knowledge, to deal with corpus-based semantic relation identification in a multiclass classification task. Few exceptions include the works by Turney \shortcite{turney2008uniform} on similarity, antonymy and analogy, and by Pantel and Pernacchiotti \shortcite{Pantel2006} on Espresso, a weakly supervised, pattern-based algorithm. Both these systems are based on patterns, which are known to be more precise than DSMs, even though they suffer from lower recall (i.e. they in fact require words to co-occur in the same sentence). DSMs, on the other hand, offer higher recall at the cost of lower precision: while they are strong in identifying distributionally similar words (i.e. nearest neighbors), they do not offer any principled way to discriminate between semantic relations (i.e. the nearest neighbors of a word are not only its synonyms, but they also include antonyms, hypernyms, and so on).

The attempts to provide DSMs with the ability of automatically identifying semantic relations include a large number of unsupervised methods \cite{weeds2003general,lenci2012identifying,santus2014taking}, which are unfortunately far from achieving the perfect accuracy. In order to achieve higher performance, supervised methods have been recently adopted, also thanks to their ease \cite{weeds2014learning,roller2014inclusive,Kruszewski2015boolean,roller2016relations,santus2016nine,nguyen2016integrating,shwartz2016improving}. Many of them rely on distributional word vectors, either concatenated or combined through algebraic functions. Others use as features either patterns or scores from the above-mentioned unsupervised methods. While these systems generally obtain high performance in classification tasks involving a single semantic relation, they have rarely been used on multiclass relation classification. On top of it, some scholars have questioned their ability to really learn semantic relations \cite{levy2015supervised}, claiming that they rather learn some lexical properties from the word vectors they are trained with. This was also confirmed by an experiment carried out by Santus et al. \shortcite{santus2016nine}, showing that up to 100\% synthetic switched pairs (i.e. \textit{banana-animal}; \textit{elephant-fruit}) are misclassified as hypernyms if the system is not provided with some of these negative examples during training.

Recently, count based vectors have been substituted by prediction-based ones, which seem to slightly improve the performance in some tasks, such as similarity estimation \cite{baroni2014predict}, even though Levy et al. \shortcite{Levy2015Hyperparameters} demonstrated that these improvements were most likely due to the optimization of hyperparameters that were instead left unoptimized in count-based models (for an overview on word embeddings, see Gladkova et al. \shortcite{Gladkova2016Analogy}). On top of it, when combined with supervised methods, the low interpretability of their dimensions makes it even harder to understand what the classifiers actually learn \cite{levy2015supervised}.

Finally, the recent attempt of Shwartz et al. \shortcite{shwartz2016improving} of combining patterns and distributional information achieved extremely promising results in hypernymy identification.

\section{System description}

Our system, ROOT18, is a Random Forest classifier \cite{breiman2001random} and it is based on the 18 features described in the following subsections. The system in its best setting makes use of the Gini impurity index as the splitting criterion and has 10 as the maximum tree depth. The half of the total number of features were considered for each split.

%Qui occorre mettere:
%1) descrizione dello spazio distribuzionale e del corpus da cui sono state estratte le frequenze
%2) descrizione delle features usate
%3) descrizione del classificatore (ed eventuali altri classificatori, nel caso li si voglia mettere per comparazione)
%4) descrizione del dataset di valutazione (Notare che il dataset è split per lexical memorization, che molte pairs hanno parole molto ambigue e senza pos: fire-shoot = SYN)

\subsection{Data}
Our data come from \textit{ukWaC} \cite{baroni2009wacky}, a 2 billion tokens corpus of English built by crawling the .uk Internet domain. For the extraction of our features, we generated several distributional spaces, which differ according to the window size  and to the statistical association measure that was used to weight raw co-occurrences. Since we obtained the best performances with window size 2 and Positive Pointwise Mutual Information \cite{church1990word}, we report the results only for this setting.
\subsection{Features}

\subparagraph{Frequency}
It is a basic property of words and it is a very discriminative information. In this type of task, it proved to be competitive in identifying the directionality of pairs of hypernyms \cite{weeds2003general}, since we expect hypernyms to have higher frequency than hyponyms. For each pair, we computed three features: the frequency of each word (\textit{Freq1,2}) and their difference (\textit{DiffFreq}).

\subparagraph{Co-occurrence}
We compute the co-occurrence frequency (\textit{Cooc}) between the two terms in each pair. This measure has been claimed to be particularly useful to spot antonyms \cite{murphy2003semantic}, since they are expected to occur in the same sentence more often than chance (e.g. \textit{Are you friend or foe?}).

\subparagraph{Entropy}
In information theory, this score is related to the informativeness of a message: the lower its entropy, the higher its informativeness \cite{Shannon1948}. Moreover, subordinate terms tend to have higher amounts of informativeness than superordinate ones. We computed the entropy of each word in the pair (\textit{Entr1,2}), plus the difference between entropies (\textit{DiffEntr}).
%Entropy is calculated according to Shannon's equation:
%$$ H(w) = -\sum_{i=1}^{n}p(c_i|w)*log_2(p(c_i|w)) $$
%where $p(c i |w)$ is the probability of the context $c_i$ given the word $w$, computed as the ratio between the co-occurrence frequency of the pair $<w, c_i>$ and the total frequency of $w$.
%Entropy corrisponds to three features in our model: one for each word in the pair (\textit{Entr1,2}), plus the difference between entropies (\textit{DiffEntr}).

\subparagraph{Cosine similarity}
It is a standard measure in DSMs to compute similarity between words \cite{turney2010frequency}. This measure is very useful to discriminate between related and unrelated terms.

$$ sim(\vec{u},\vec{v}) = \frac{\vec{u} \cdot \vec{v}}{ \| \vec{u} \| \cdot  \| \vec{v} \|} $$

\subparagraph{LinSimilarity}
LinSimilarity \cite{lin1998information} is a different similarity measure, computed as the ratio of shared context between \textit{u} and \textit{v} to the contexts of each word:
$$ Lin(\vec{u},\vec{v}) = \frac{\sum _{c \in \vec{u} \bigcap \vec{v}}[\vec{u}[c]+\vec{v}[c]}{\sum_{c \in \vec{u}} \vec{u}[c] + \sum_{c \in \vec{v}} \vec{v}[c]} $$

\subparagraph{Directional similarity measures} 
We extracted several directional similarity measures that were proposed to detect hypernyms, such as \textit{WeedsPrec}, \textit{cosWeeds}, \textit{ClarkeDe} and \textit{invCL} (for a review, see Lenci and Benotto  \shortcite{lenci2012identifying}).
They are all based on the \textit{Distributional Inclusion Hypotesis}, according to which if a word \textit{u} is semantically narrower to \textit{v}, then a significant number of the salient features of \textit{u} will be included also in \textit{v}.

\subparagraph{APSyn}
This measure and the following \textit{APAnt} do not rely on the full distribution of words, but on the top \textit{N} most related contexts of the words according to some statistical association measure.
APSyn \cite{santus2016what} computes a weighted intersection of the top \textit{N} context of the target words:
$$
APSyn(w_1,w_2) = \sum_{f \in N(F_1) \bigcap N(F_2)} \frac{1}{(rank_1(f)+rank_2(f))/2}
$$
That is, for every feature \textit{f} included in the intersection between the top N features of $w_1$ and $w_2$ ($N(F_1)$, $N(F_2)$ respectively), the measure adds 1 divided by the average rank of the feature in the rankings of the top N features of $w_1$ and $w_2$.
%, among the top ranked features of $w_1$, $rank_1(f_1)$, and $w_2$, $rank_2(f_2)$.

%For each pair, we computed APSyn for the top 1000 and for the top 100 contexts.

\subparagraph{APAnt} 
\textit{APAnt} \cite{santus2014taking} is defined as the inverse of APSyn. This unsupervised measure tries to discriminate between synonyms and antonyms by relying on the hypothesis that words with similar distribution (i.e. high vector cosine) that do not share their most relevant contexts (i.e. what APSyn computes) are likely to be antonyms. For each pair, we computed APSyn and APAnt for the top 1000 and for the top 100 contexts.

%Average Precision to estimate the extent and salience of the intersection between the N most salient contexts of terms: the broader and more salient the intersection, the higher the probability they are synonyms; \textit{vice versa}, they are antonyms. It is formally defined as:
%$$
%APAnt(w_1,w_2) = 1 / \sum_{f \in F_1 \bigcap F_2} \frac{1}{min(rank_1(f_1)+rank_2(f_2))}
%$$
%where $F_x$ is the set of the N most salient features of $x$ and $rank_x(f_x)$ is the rank of feauture $f_x$ in the salience ranked feature list for the term $x$.

\subparagraph{Same POS}

We realized that many of the random pairs in the data included words with different parts of speech. Therefore, we decided to add a boolean value to our set of features: 1 if the most frequent POS of the words in the pair were the same, 0 otherwise.

%\subsection{Classifier}

%ROOT18 is a Random Forest classifier \cite{breiman2001random}, and it is based on the 18 features described above. We report the results for the system in its best settings, i.e. the Gini impurity index as the splitting criterion and 10 as the maximum depth of the trees. The half of the total number of the features were considered for each split.

\subsection{Evaluation dataset}

The task organizers provided a training and a test set extracted from EVALution 1.0, a resource that was specifically designed for evaluating systems on the identification of semantic relations \cite{santus2015evalution}. EVALution 1.0 was derived from WordNet \cite{fellbaum1998wordnet} and ConceptNet \cite{liu2004concept} and it consists of almost 7500 word pairs, instantiating several semantic relations.

The training and the test set included, respectively, 3054 and 4260 word pairs and they are lexical-split, that is, the two sets do not share any pair. Since words were not tagged, we performed POS-tagging with the TreeTagger \cite{schmid1995tree}.

\section{Results}

\begin{table}[ht]
\small
\centering
\label{my-label1}
\begin{tabular}{|c|c|c|c|c|c|c|}
\hline
	\textbf{Model} & \textbf{P (task1)} & \textbf{R (task1)} & \textbf{F (task1)} & \textbf{P (task2)} & \textbf{R (task2)} & \textbf{F (task2)} \\ \hline
	Random Baseline & 0.283 & 0.503 & 0.362 & 0.073 & 0.201 & 0.106 \\ \hline
    Cosine Baseline & 0.589 & 0.573 & 0.581 & 0.170 & 0.165 & 0.167 \\ \hline
    ROOT18(100) & 0.818 & 0.657 & 0.729 & 0.304 & 0.213 & 0.249	\\ \hline
	ROOT18(500) & 0.818 & 0.650 & 0.724 & 0.313 & \textbf{0.227} & \textbf{0.262}  \\ \hline
	ROOT18(1000) & \textbf{0.823} & \textbf{0.657} & \textbf{0.731} & \textbf{0.343} & 0.218 & 0.261  \\ \hline
\end{tabular}
\caption{Precision, Recall and F-measure scores for subtask 1 and 2. The numbers between parentheses in the ROOT18 rows refer to the number of estimators used by the classifier.} 
\end{table}

%\begin{table}[ht]
%\small
%\centering
%\label{my-label2}
%\begin{tabular}{|c|c|c|c|}
%\hline
%	\textbf{Model} & \textbf{Precision} & \textbf{Recall} & \textbf{F-measure} \\ \hline
 %   Random baseline & 0.073 & 0.201 & 0.106 \\ \hline
 %   Cosine baseline & 0.170 & 0.165 & 0.167 \\ \hline
%	ROOT18, 100 estimators & 0.304 & 0.213 & 0.249	\\ \hline
%	ROOT18, 500 estimators  & 0.313 & \textbf{0.227} & \textbf{0.262}  \\ \hline
%	ROOT18, 1000 estimators & \textbf{0.343} & 0.218 & 0.261  \\ \hline
%\end{tabular}
%\caption{Precision, recall and F-measure scores for subtask 2.} 
%\end{table}

As it can be seen from table 1, ROOT18 has a solid performance on the subtask 1, and it is quite accurate in separating related terms from unrelated ones. Generally speaking, we noticed that the classifier performs better when Gini impurity index is used as a splitting criterion instead of entropy. % and with a higher number of estimators. ????????????? LA FRASE NON E' CHIARA
The model with 1000 estimators is our best performing one, with Precision = 0.823, Recall = 0.657 and F-score = 0.731. Concerning the contribution of the features, APSyn1000 and vector cosine have the highest relative importance, with respective contributions of 0.29 and 0.12 to the prediction function. This is not at all surprising, since APSyn and cosine already proved to be strong predictors of semantic similarity.

\begin{table}[ht]
\small
\centering
\label{my-label3}
\begin{tabular}{|c|c|c|c|}
\hline
	\textbf{Relation} & \textbf{Precision} & \textbf{Recall} & \textbf{F-measure} \\ \hline
    SYN & 0.309 & 0.179 & 0.226 \\ \hline
	ANT & 0.298 & 0.206 & 0.243	\\ \hline
	HYPER  & 0.397 & 0.343 & 0.368  \\ \hline
	PART-OF & 0.200 & 0.116 & 0.147  \\ \hline
\end{tabular}
\caption{Precision, recall and F-measure for each relation in subtask 2 (ROOT-18 with 500 estimators).} 
\end{table}
\begin{table}[h]
\small
\centering
\label{my-label4}
\begin{tabular}{|c|c|c|c|c|c|}
\hline
	Relation & SYN & ANT & HYPER & PART-OF & RANDOM \\ \hline
    SYN & 42 & 29 & 58 & 24 & 82 \\ \hline
	ANT & 29 & 74 & 38 & 23 & 196	\\ \hline
	HYPER  & 32 & 46 & 131 & 30 & 143  \\ \hline
	PART-OF & 15 & 43 & 59 & 26 & 81 \\ \hline
    RANDOM & 18 & 56 & 44 & 27 & 2914  \\ \hline
\end{tabular}
\caption{Confusion matrix for subtask 2 (ROOT-18 with 500 estimators).} 
\end{table}

Results are much less convincing for subtask 2. In particular, the recall values are extremely low, especially for some of the semantic relations: part\_of, for example, is often below 0.15. For such relation we have no dedicated features in our system, so the difficulty in identifying meronyms are not a surprise.
On the other hand, ROOT18 showed the benefits of the inclusion of several measures targeting hypernymy, since the latter is the most accurately recognized relation (precision often $>$ 0.4), recording also the higher recall (always $>$ 0.3, even in the worst performing models).

The performance did not show any particular improvement by increasing the number of the decision trees, so that our best overall results are obtained by the model with 500 estimators (precision = 0.343, recall = 0.218 and F-score = 0.261). As for the contributions of the single features, APSyn1000 (0.19) and cosine (0.09) are still the top ones, followed by cosWeeds (0.07) and APAnt1000 (0.06).

Table 4 describes the confusion matrix, which shows that randoms are properly working as distractors for the model, leading to a large number of misclassification. Synonyms are often confused with hypernyms and this might be due to the fact that the difference between the two is subtle. These results suggest that measures based on the Distributional Inclusion Hypothesis are not always efficient in discriminating between synonyms and hypernyms. 

Antonyms are confused with hypernyms and \textit{vice versa}, which might be due to the fact that neither share their most relevant features, obtaining therefore similar APAnt scores \cite{santus2015when}. Meronyms, finally, are mostly confused with hypernyms, which is almost surely due to the generality spread that characterize both relations and that is captured by both frequency and entropy in our system.

\subsection{Conclusions}

Our results clearly highlight the difficulty of DSMs in discriminating between several semantic relations at once. Such models, in fact, rely on a vague definition of semantic similarity (i.e. distributional similarity) which does not offer any principled way to distinguish among different types of semantic relations. 

Nonetheless, it is still feasible for traditional DSMs to achieve good performances on the recognition of taxonomical relations \cite{santus2016nine}, for which metrics can be defined on the basis of feature inclusion, of context informativeness etc.  For other relations, such as antonymy and meronymy, it is not easy to define measures based on distributional similarity (for the latter relation, it is difficult even to find an univocal definition: see Morlane-Hond{\`e}re \shortcite{morlanehondere2015what}): APAnt works relatively well in discriminating antonyms from synonyms, but -- as noticed by Santus et al. \shortcite{santus2015when} -- this measure has also a bias towards hypernyms, which explains why these are often confused. A possible solution, in our view, would be the integration of DSMs with pattern-based information, in a way that is already being shown by some of the current state-of-the-art systems (see, for example, Shwartz et al. \shortcite{shwartz2016improving}). Such integration has the advantage of combining the precision of the patterns with the high recall of DSMs.

Finally, we may assume that also the configuration of the original dataset could contribute to our results, since some pairs in the dataset have ambiguous words and the target relations hold for only one of the their meanings. Disambiguating the pairs, at least by Part-Of-Speech, would certainly help in improving the results. A simple method might consist in computing the vector cosine for the pairs with the target words declined in all possible POS (i.e. VV, NN, JJ) and then maintain in the dataset only the pair with the higher value.

\section{Acknowledgements}

This work has been carried out thanks to the support of the A*MIDEX grant (n ANR-11-IDEX-0001-02) funded by the French Government "Investissements d'Avenir" program.

%\bibliography{coling2016}

\end{document}